\documentclass[conference]{./support/ieeeconf}
\usepackage{times}
\usepackage{amsmath}
\usepackage[pdftex]{graphicx}
\usepackage{subfigure}
\usepackage{amsmath,amssymb,amsopn,amstext,amsfonts}
\usepackage{cancel}
\usepackage[space]{cite}
\usepackage{pdfsync}
\usepackage{balance}
\usepackage{color}
\usepackage{algorithm}
\usepackage{algorithmic}
\usepackage{url}
\usepackage{booktabs}
\usepackage{threeparttable}
\usepackage[linkcolor=black,citecolor=black,urlcolor=black,colorlinks=true]{hyperref}
\usepackage{multirow}
\usepackage[outdir=./epstopdf/]{epstopdf}
\usepackage{graphicx}
\IEEEoverridecommandlockouts

\DeclareMathOperator*{\argminA}{arg\,min}

\graphicspath{{./figure/}}
\DeclareGraphicsExtensions{.pdf,.png,.jpg,.eps}
\title{\LARGE \bf AVP-SLAM: Semantic Visual Mapping and Localization \\ for Autonomous Vehicles in the Parking Lot}
\author{Tong Qin, Tongqing Chen, Yilun Chen, and Qing Su
\thanks{All authors are with  IAS BU, Huawei Technologies, Shanghai, China.
        {\tt\small \{qintong, chentongqing, chengyilun, suqing\}@huawei.com}.}
}

\begin{document}

\maketitle
\thispagestyle{empty}
\pagestyle{empty}

\begin{abstract}

Autonomous valet parking is a specific application for autonomous vehicles.
In this task, vehicles need to navigate in narrow, crowded and GPS-denied parking lots. 
Accurate localization ability is of great importance.
Traditional visual-based methods suffer from tracking lost due to texture-less regions, repeated structures, and appearance changes.
In this paper, we exploit robust semantic features to build the map and localize vehicles in parking lots.
Semantic features contain guide signs, parking lines, speed bumps, etc, which typically appear in parking lots.
Compared with traditional features, these semantic features are long-term stable and robust to the perspective and illumination change.
We adopt four surround-view cameras to increase the perception range. 
Assisting by an IMU (Inertial Measurement Unit) and wheel encoders, 
the proposed system generates a global visual semantic map.
This map is further used to localize vehicles at the centimeter level. 
We analyze the accuracy and recall of our system and compare it against other methods in real experiments.
Furthermore, we demonstrate the practicability of the proposed system by the autonomous parking application.

\end{abstract}


\section{Introduction}
There is an increasing demand for autonomous driving in recent years.
Accurate localization is the most important prerequisite for autonomous applications.
Perception, prediction, planning, and control are all based on localization results.
To achieve robust localization, vehicles are equipped with various sensors, such as GPS, camera, Lidar, IMU, wheel odometer, etc.
A great number of localization methods appeared over the last decades, such as visual-based methods \cite{mur2017orb, engel2014lsd}, visual-inertial-based methods \cite{qin2018vins, LeuFurRab1306, MouRou0704}, Lidar-based methods \cite{zhang2014loam, le2019in2lama, yin20203d}.
For commercial-level production, low-cost sensors, such as IMUs and cameras, are preferred.
Besides localization, mapping is also an important capability for autonomous drivings.
For some private areas, such as closed factory parks, parking lots, there are no predefined maps.
Vehicles need to build the map by itself. 
So localization and mapping abilities are of great importance for autonomous driving.

\begin{figure}
    \vspace{1.0cm}
    \centering
    \includegraphics[width=0.48\textwidth]{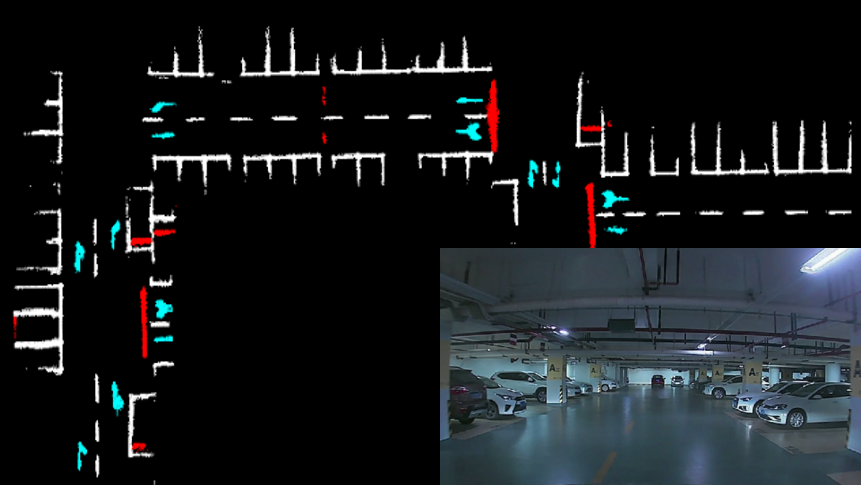}
    \caption{The figure in the right corner shows a common scene in the underground parking lot, where the environment is narrow, light-less, and GPS-denied. Autonomous driving in such environment is challenging. The bigger figure is the semantic visual map of this parking lot, which consists of semantic features (guide signs, parking lines, and speed bumps).
    This map can be used to localize vehicles at centimeter-level accuracy.
    Video Link:	\url{https://youtu.be/0Ow0U-G7klM}}
    \label{fig:abs}
\end{figure}

\begin{figure*}
    \centering
    \includegraphics[width=0.9\textwidth]{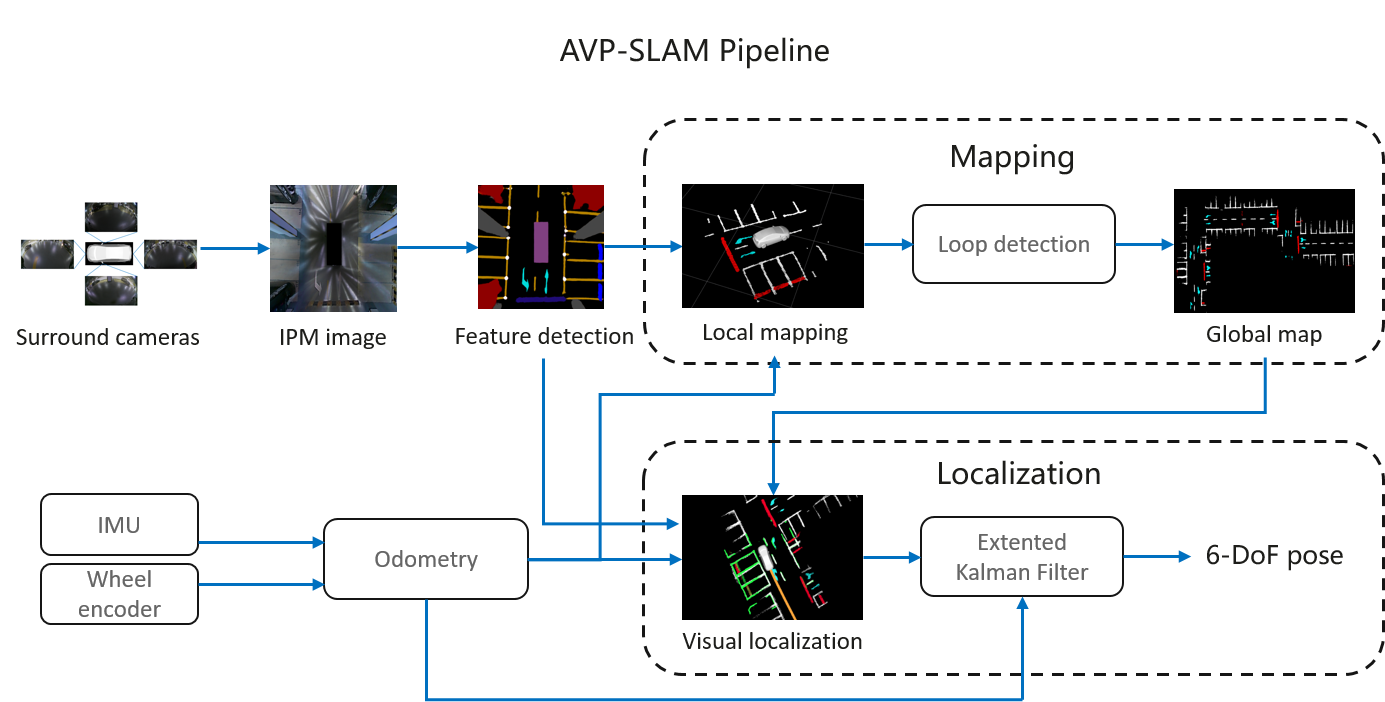}
    \caption{Block diagram illustrating the full pipeline of the proposed AVP-SLAM system. The system starts with four surround cameras, an IMU and wheel encoders. In the mapping procedure, it builds a map of the environment with semantic features. Based on this prior map, the 6-DoF pose can be localized in the localization procedure at centimeter-level accuracy.}
    \label{fig:framework}
\end{figure*}

Autonomous valet parking is a specific application for autonomous driving, where vehicles need to navigate into parking lots and park into the target spot automatically.
Since parking lots are usually small, narrow, and crowded, precise localization is of crucial importance.
Any localization error may cause a crash unexpectedly.
To avoid the high cost of sensors, studies have focused on vision-based localization methods.
However, this scenario puts great challenges to traditional visual localization methods. 
On one hand, indoor and underground parking lots are mostly made up of texture-less walls, poles, and grounds.
Feature detection and matching are unstable.
Traditional visual methods easily suffer from tracking lost.
On the other hand, different vehicles may stay in various parking places on different days, which makes the appearance change a lot.
It's impossible to relocalize vehicles by appearance-based maps in a long time.

To solve this problem, we adopt a new kind of feature, semantic feature.
Semantic features contain guide signs, parking lines, and speed bumps, which typically appear in parking lots.
Compared with traditional geometrical features, these semantic features are long-term stable and robust to perspective and illumination changes.
In this work, we proposed a semantic-feature-based mapping and localization system, which enables vehicles to automatically navigate in parking lots. The contribution of this paper is summarized as follows:
\begin{itemize}
    \item We propose a novel type of semantic features used in the visual SLAM framework.
    \item We propose a complete mapping and localization system for autonomous driving in the parking lot.
    \item We conduct the real-world autonomous parking application based on the proposed system.
\end{itemize}








\section{literature review}
There is a great number of research works related to visual-based localization over the last decades.
Based on output types, we divided them into relative localization and global localization.
Meanwhile, based on the feature types, we category them into traditional methods and road-based methods.

\textbf{Relative localization v.s. Global localization}
Relative localization is also called as odometry, which initializes coordinate at the start position, and focuses on the relative pose between local frames. 
Popular relative localization includes visual odometry \cite{klein2007parallel, ForPizSca1405, engel2014lsd, mur2017orb, kitt2010visual}, visual-inertial odometry \cite{MouRou0704, qin2018vins, LeuFurRab1306, forster2017manifold}, Lidar odometry \cite{zhang2014loam, le2019in2lama, yin20203d}.
On the contrary, the global localization has a fixed coordinate.
It usually localizes against a prior map.
For example, visual-based methods \cite{lynen2015get, burki2016appearance} localized camera pose against a visual feature map by descriptor matching. 
The map contains thousands of 3D visual features and their descriptors.   
Methods, like \cite{schneider2018maplab, QinShen18}, can automatically merge multiple sequences into a global map.
HD (High Definition) maps, which contain accurate road elements, are often used for global localization in autonomous driving area \cite{schreiber2013laneloc, ranganathan2013light, lu2017monocular}.

\textbf{Traditional feature methods} exploit geometrical features such as sparse points, lines, and dense planes in the natural environment. 
Corner feature points are widely used in visual odometry algorithms \cite{klein2007parallel, mur2017orb, qin2018vins, LeuFurRab1306, LiMou1305}.
Camera pose, as long as feature positions, are estimated in these algorithms.
Features can be further described by surround patch to make it distinguishable, such as SFIT, SURF, ORB, BRIEF descriptor, etc.
More than visual odometry, these methods, \cite{lynen2015get, mur2017orb, schneider2018maplab, burki2016appearance, QinShen18}, can build a visual map in advance, then relocalize camera pose within this map.
For example, Mur-Artal \cite{mur2017orb} leveraged ORB features to build the environment map.
Then the map can be used to relocalize the camera by ORB descriptor matching the next time.
For autonomous driving tasks, Burki \cite{burki2016appearance} demonstrated vehicles were localized by the sparse feature map on the road.
Inherently, traditional feature-based methods were suffered from lighting, perspective, and appearance changes in the long term. 

\textbf{Road-based feature methods} adopt land markings on the road surface, which are widely applied to autonomous driving applications.
Land markings include lane lines, curbs, makers, etc.
These methods localized camera pose by matching land markings with a prior map.
Compared with traditional features, these markings are more robust against illumination changes and stable in the long term.
For global localization, an accurate prior map is necessary.
This prior map is usually built by other extended sensor setups (Lidar, GNSS, etc.).
Schreiber \cite{schreiber2013laneloc} localized camera by detecting curbs and lanes, and matching features with a highly accurate map.
Further on, Ranganathan \cite{ranganathan2013light} detected corner points on road markings, and used corner points to perform localization.
These corner points were more distinct.
Yan \cite{lu2017monocular} not only matched the geometry of road markings but also took vehicle odometry and epipolar geometry constraints into account and formulated a non-linear optimization problem to estimate the 6 DoF camera pose.
Meanwhile, some research focused on building a road map.
Regder \cite{rehder2015submap} detected lanes on image and used odometry to generate local grid maps and optimized pose by local map stitching.
Moreover, Jeong \cite{jeong2017road} classified road marking and only incorporated informative classes to avoid ambiguity.
He eliminated accumulated drift by loop closure and maintained the consistency by the pose graph optimization.

In this paper, we deal with a more challenging scenario than aforementioned methods.
That is the underground parking lot, where the environment is narrow, light-less, GPS-denied, and without prior maps.
We adopt a novel semantic feature for localization and mapping. 
Semantic features (guide signs, parking lines, and speed bumps) are detected by a convolution neural network.
A global semantic map is generated from the mapping procedure.
Then this map is used to localize vehicles at centimeter-level accuracy.

\begin{figure}
    \centering
    \includegraphics[width=0.4\textwidth]{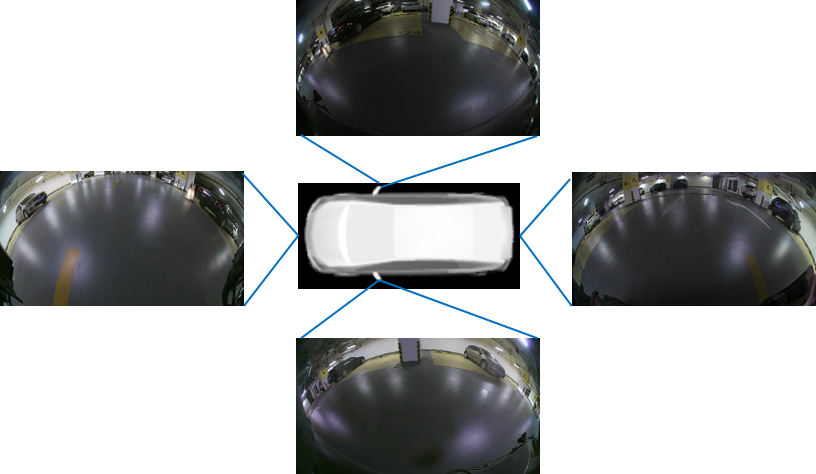}
    \caption{Illustration of the configuration of four surround-view cameras used in AVP-SLAM.}
    \label{fig:surround_camera}
\end{figure}

\section{System Overview}
The system adopts four surround-view cameras to increase perception range.
This is a common setup for high-configuration cars nowadays.
The system also uses an IMU (Inertial Measurement Unit) and two wheel encoders.
The IMU and wheel encoders form odometry, which provides relative pose but suffers from accumulative error.
Intrinsic and extrinsic parameters of all sensors were calibrated offline in advance.

The framework consists of two parts, as shown in Fig. \ref{fig:framework}.
The first part is mapping, which builds a global semantic map for the environment.
The four images from surround cameras are projected into the bird's eye view and synthesized into one omnidirectional image.
The neural network detects semantic features, which include lanes, parking lines, guide signs, and speed bumps.
Based on the odometry, semantic features are projected into a global coordinate.
Since the odometry drifts in the long run, we detect loop closure by local map matching to reduce accumulated error.
The second part is the localization. 
As same as the mapping part, semantic features are extracted from the bird's eye image. 
Vehicles are localized by matching semantic features with a previous-build map.
In the end, an EKF (Extended Kalman Filter) fuses visual localization results with odometry, which guarantees the system has a smooth output and survives in the texture-less region.

\section{methodology}

\begin{figure}
    \centering
    \subfigure[The IPM image.]{
        \label{fig:ipm}    
        \includegraphics[width=0.47\columnwidth]{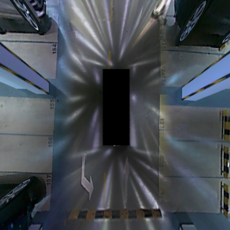}}   
    \subfigure[The segmentation image.]{
        \label{fig:semantic}
        \includegraphics[width=0.47\columnwidth]{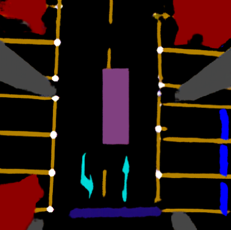}}
    \caption{IPM image segmentation. (a) is the synthetic IPM image from four surround-view cameras. (b) is the segmentation result of this IPM image. Red area denotes obstacles. Orange area denotes lanes and parking lines. Grey area denotes walls. Blue area denotes speed bumps. Cyan area denotes guide signs. White dots denote corners of parking spot.}
    \label{fig:ipm_semantic_feature}
\end{figure}

\subsection{IPM Image}

Four surround-view cameras are used in AVP-SLAM.
The camera configuration is shown in Fig. \ref{fig:surround_camera}, which is a common setup for the high-configuration commercial cars.
One camera is in the front, one camera is in the rear, and two cameras are on the left and right separately.
These cameras are equipped with fisheye lens and look downward.

The intrinsic and extrinsic parameters of each camera are calibrated offline.
Each pixel is projected into the ground plane (z equals 0) under the vehicle center coordinate, which is also called IPM (Inverse Perspective Mapping).
The projection process is conducted as follows,

\begin{equation}
\label{eq:ipm}
\frac{1}{\lambda}
\begin{bmatrix}
x^v \\ y^v \\ 1
\end{bmatrix}
= 
\begin{bmatrix}
\mathbf{R}_c \ \mathbf{t}_c
\end{bmatrix}
_{col:1,2,4} ^ {-1}
\pi_c ^ {-1}(
\begin{bmatrix}
u\\v\\1
\end{bmatrix}
),
\end{equation}
where $\pi_c(\cdot)$ is the projection model of fisheye.
$\pi_c(\cdot)^{-1}$ is the inverse projection, which lifts pixel into space. 
$[\mathbf{R}_c \ \mathbf{t}_c]$ is the extrinsic matrix of each camera with respect to vehicle's center.
$[u \ v]$ is pixel location in image coordinate.
$[x^v \ y^v]$ is the position of the feature in the vehicle's center coordinate.  
$\lambda$ is a scalar.
$()_{col:i}$ means taking the $i_{th}$ column of this matrix.

After the inverse perspective projection, we synthesize points from four images into a big one, as shown in Fig. \ref{fig:ipm}. 
The reprojection is carried out by the following equation:
\begin{equation}
\label{eq:ipm2}
\begin{bmatrix}
u_{ipm} \\ v_{ipm} \\ 1
\end{bmatrix}
=
\mathbf{K}_{ipm}
\begin{bmatrix}
x^v\\y^v\\1
\end{bmatrix}
,
\end{equation}
where $\mathbf{K}_{ipm}$ is the intrinsic parameter of the synthesized IPM image.
$[u_{ipm} \ v_{ipm}]$ is pixel location in the synthesized IPM image.

This synthesized image contains omni-directional information, which dramatically increases the perception range.
It is very useful in the narrow parking lot where occlusion happens a lot. 

\subsection{Feature Detection}
We adopt the popular CNN (Convolution Neural Network) method for semantic feature detection.
A lot of segmentation networks can be used for feature detection, such as \cite{long2015fully, ronneberger2015u, badrinarayanan2015segnet}.
In this paper, we modified U-Net\cite{ronneberger2015u} to segment images into different categories.
This network is specially trained with images captured in parking lots, which classifies pixels into lanes, parking lines, guide signs, speed bumps, free space, obstacles, and walls.
A sample of results is shown in Fig. \ref{fig:ipm_semantic_feature}.
The input is an IPM image, Fig. \ref{fig:ipm}.
The segmentation results are drawn in Fig. \ref{fig:semantic}.
Among these classes, parking lines, guide signs, and speed bumps are distinct and stable features, which are used for localization.
The parking line is also used for parking spot detection. 
Free space and obstacles are used for planning.


\subsection{Local Mapping}

After image segmentation, useful features (parking lines, guide signs, and speed bumps) are lifted into 3D space as follows 
,

\begin{equation}
\label{eq:lift}
\begin{bmatrix}
x^v\\y^v\\1
\end{bmatrix}
=
\mathbf{K}_{ipm}^{-1}
\begin{bmatrix}
u_{ipm} \\ v_{ipm} \\ 1
\end{bmatrix}
.
\end{equation}

Based on the odometry, the features are transferred from vehicle coordinate into the world coordinate as follows,

\begin{equation}
\label{eq:trans}
\begin{bmatrix}
x^w \\ y^w \\ z^w
\end{bmatrix}
= \mathbf{R_o} 
\begin{bmatrix}
x^v \\ y^v \\ 0
\end{bmatrix}
+ \mathbf{t}_o
,
\end{equation}
where $[\mathbf{R_o} \ \mathbf{t}_o]$ is the pose from odometry.
These points aggregates into a local map. 
We maintain a local map for every 30 meters.
The sample of local maps is shown in Fig. \ref{fig:loop1} and \ref{fig:loop2}.

\subsection{Loop Detection}
Since odometry drifts in a long time, we detect loop closure to eliminate drift.
For the latest local map, we compare it with other surround local maps.
Two local maps are matched by the ICP (Iterative Closest Point) method.
If two local maps match successfully, we get the relative pose between these two local maps.
This relative pose will be used in the global pose graph optimization to correct drift.

\begin{figure}
    \centering
    \subfigure[Local map I]{
        \label{fig:loop1}    
        \includegraphics[width=0.47\columnwidth]{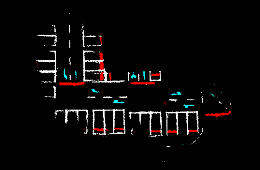}}   
    \subfigure[Local map II]{
        \label{fig:loop2}
        \includegraphics[width=0.47\columnwidth]{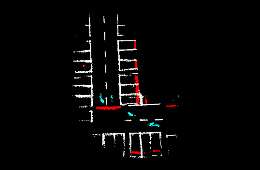}}
        \subfigure[Local map I \& II merging]{
        \label{fig:loop3}
        \includegraphics[width=0.47\columnwidth]{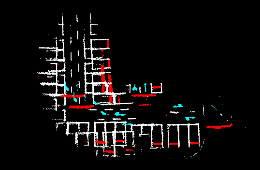}}
        \subfigure[Local map I \& II merging after loop detection]{
        \label{fig:loop4}
        \includegraphics[width=0.47\columnwidth]{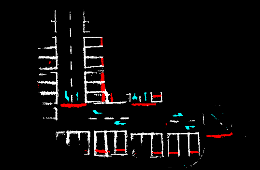}}
    \caption{Illustration of the loop detection procedure. Fig. (a) and (b) show two local map separately. These two local map are the same place visited at different times instants. Directly merging two local maps, we get results shown in Fig. (c). Because of drifts, two maps don't match well. Fig. (d) shows perfect matching result after loop detection.}
    \label{fig:loop}
\end{figure}

For example, Fig. \ref{fig:loop1} and Fig. \ref{fig:loop2} are two local maps.
If we directly merge two local maps, these two local maps overlap badly due to the odometry drift, as shown in Fig. \ref{fig:loop3}.
By loop detection, we find the relative pose between two local maps, so that these local maps match perfectly, as shown in Fig. \ref{fig:loop4}.

\subsection{Global Optimization}

After loop detection happens, a global pose graph optimization is performed to eliminate accumulated drift and maintain the consistency of the whole graph.
In this pose graph, node is the pose of every local map, which contains three axis rotations, $\mathbf{r} = [r_x \ r_y \ r_z]^T$, and translation, $\mathbf{t} = [t_x \ t_y \ t_z]^T$.
There are two kinds of edges.
One is the odometry edge, which constrains two sequential local maps by odometry measurements.
Another one is the loop closure edge, which constrains looped local maps.
The pose graph optimization can be formulated as following cost function,

\begin{equation}
\begin{split}
\mathcal{X}^* =
 \argminA_{\mathcal{X}} & \sum_t \| f(\mathbf{r}_{t+1},\mathbf{t}_{t+1},\mathbf{r}_t,\mathbf{t}_t ) - \mathbf{z}^o_{t, t+1}  \|^2 \\
+ & \sum_{{i,j}\in\mathcal{L}} \| 
f(\mathbf{r}_{i},\mathbf{t}_{i},\mathbf{r}_j,\mathbf{t}_j )
- \mathbf{z}^l_{i,j}\|^2 
\end{split}
,
\end{equation}
where $\mathcal{X} = [\mathbf{r}_0, \mathbf{t}_0, ..., \mathbf{r}_t, \mathbf{t}_t]^T$, which is poses of all local maps.
$\mathbf{z}^o_{t, t+1}$ is the relative pose between local map $t$ and $t+1$ from odometry.
$\mathcal{L}$ is the set of all looped pairs.
$\mathbf{z}^l_{i,j}$ is the relative pose between loop frame $i$ and $j$.
The function $f(\cdot)$ computes relative pose between two local maps.
The optimization is carried out by the Gauss-Newton method.

After global pose graph optimization, we stack local maps together by updated poses.
In this way, a globally consistent map is generated.

\subsection{Localization}

\begin{figure}
    \centering
    \includegraphics[width=0.4\textwidth]{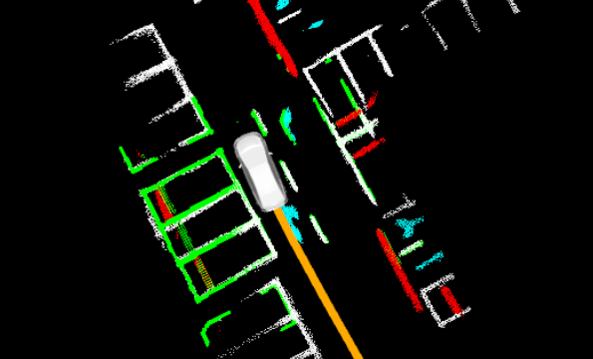}
    \caption{Illustration of localization in semantic map. The white, red, and blue dots are parking lines, speed bump, guide signs in the map. The green dots are current features. The orange line is the estimated trajectory. The vehicle is localized by matching current features with the map.}
    \label{fig:reloc}
\end{figure}

Based on this semantic map, the vehicle can be localized when it comes to this parking lot again.
Similar to the mapping procedure, surround-view images are synthesized into one IPM image.
Semantic features are detected on the IPM image and lifted into the vehicle coordinate.
Then the current pose of the vehicle is estimated by matching current feature points with the map, as shown in Fig. \ref{fig:reloc}.
The estimation adopts the ICP method, which can be written as following equation,  
\begin{equation}
\mathbf{r}^*, \mathbf{t}^* = \argminA_{\mathbf{r},\mathbf{t}} \sum_{k\in \mathcal{S}} \| \mathbf{R}(\mathbf{r}) 
\begin{bmatrix}
x^v_k \\ y^v_k \\ 0
\end{bmatrix}
 + \mathbf{t}
-
\begin{bmatrix}
x^w_k \\ y^w_k \\ z^w_k
\end{bmatrix}
 \| ^2
 ,
\end{equation}
where $\mathbf{r}$ and $\mathbf{t}$ are three dimensional rotation and translation vector of current frame.
$\mathcal{S}$ is the set of current feature points.
$[x^v_k \ y^v_k \ 0]$ is current feature under vehicle coordinate.
$[x^w_k \ y^w_k \ z^w_k]$ is the closest point of this feature in the map under global coordinate.

It is important to note that a good initial guess is critical for the ICP method.
At the very beginning, there are two strategies for initialization.
One way is that the entrance of the parking lot is marked on the map.
So the vehicle is directly initialized at the entrance of the parking lot.
The second way is that we can use GPS as an initial pose before entering the underground parking lot. 
The GPS is not used anymore after the vehicle is localized on the map. 
After initialization, the prediction from odometry is used as the initial guess. 

The localization is accurate in the textured region.
Although the surround views increase the perception range as much as possible, there exists some extremely texture-less area in the parking lot. 
To overcome this problem, an EKF framework is adopted at the end, which fuses odometry with visual localization results.
In this filter, the odometry is used for prediction, and visual localization results are used for updating.
The filter not only increases the robustness of the system but also smooths the estimated trajectory.

\subsection{Parking Spot Detection} 
\begin{figure}
    \centering
    \includegraphics[width=0.4\textwidth]{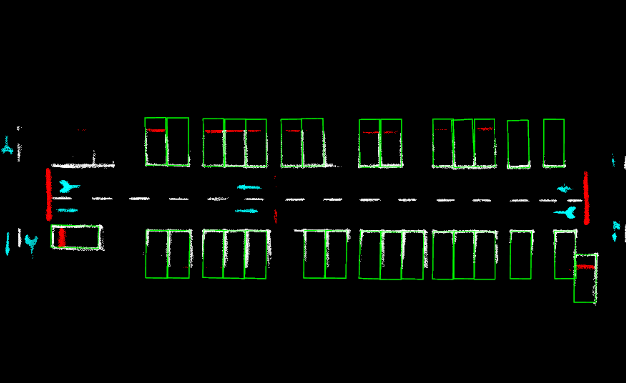}
    \caption{Parking spots are generated automatically by parking corners and parking line fitting.}
    \label{fig:parkingspots}
\end{figure}

Since parking lines and parking line corners are detected from the IPM image (white dots and yellow lines in Fig. \ref{fig:semantic}), it is easy to detect parking spots automatically.
Corners are used to predict positions of parking spots.
If the parking lines match predicted parking spots well, this prediction is considered correct. 
The result of parking spot detection is shown in Fig. \ref{fig:parkingspots}. 
These parking spots are marked on the map, which are used for autonomous parking tasks.

\section{Experimental Results}

We present experiments for validating the proposed AVP-SLAM system.
All the data used in experiments was obtained from a vehicle platform.
The vehicle was equipped with the surround-view camera system,
which consisted of four cameras with fish-eye lens, mounting at the front, rear, left, and right side respectively.
Images were recorded at 30Hz with a resolution of 1280x720 pixels.
Furthermore, wheel encoders and an IMU provided odometry measurements.
Segmentation was running in real-time at 15Hz on a consumer-grade computer.
For the metric evaluation, we used the RTK-GPS as ground truth in the open outdoor area.

Due to the uniqueness of our sensor configuration, it is hard to directly compare against other existing algorithms with the same sensor setup.
We adopted two additional front cameras to run a visual algorithm, ORB-SLAM2\cite{mur2017orb}.
We compared the proposed AVP-SLAM with ORB-SLAM2 in terms of mapping accuracy (Sect. \ref{subsec:metric}), localization accuracy, and recall rate (Sect. \ref{subsec:recall}).
AVP-SLAM achieved 1.33\% mapping error and centimeter-level localization accuracy.

\subsection{Mapping Metric Evaluation}
\label{subsec:metric}

For mapping metric evaluation, we choose an outdoor parking lot, where the GPS signal is in good quality.
The vehicle traversed along a square trajectory and backed to the starting point.
So loop closure happened in this scenario.
The total length is $324m$.
We compared proposed AVP-SLAM with pure odometry from the IMU and wheel encoder, and ORB-SLAM2.
RTK-GPS was treated as ground truth.

The absolute trajectory error is shown in Table \ref{tab:metric}.
It can be seen that pure odometry was much worse than visual methods.
Due to the measurement noise, the odometry drifted inevitably.
Both AVP-SLAM and ORB-SLAM2 compensated accumulative error by accurate loop detection using visual features. 
The difference is that one used semantic features, the other one used geometric features.
Our AVP-SLAM was a little bit better than ORB-SLAM2.
This should be contributed to more sensors we used (two additional cameras, an IMU, and two wheel encoders).

\begin{table}[h]
    \centering
    \caption{{Absolute Trajectory Error in Mapping} \label{tab:metric}}
    \begin{tabular}{c|ccc}
        \toprule
        Method & RMSE [m] & Max [m] & NEES  \\
        \midrule
        Odometry    &  7.24  & 15.53 & 2.23 \%  \\
        ORB-SLAM2   &  4.57  & 8.58  & 1.41 \%  \\
        AVP-SLAM    &  \textbf{4.31}  & \textbf{8.32}  & \textbf{1.33} \%  \\
        \bottomrule
    \end{tabular}
    \begin{tablenotes}
        \footnotesize
        \item[1] *RMSE is the root mean square error. NEES is the normalized estimation error squared, which equals to RMSE / total length.
    \end{tablenotes}
\end{table}

\subsection{Undergroud Parking Lot}
\label{subsec:recall}

\begin{figure}
    \centering
    \subfigure[The light change]{
        \label{fig:c1}    
        \includegraphics[width=0.31\columnwidth]{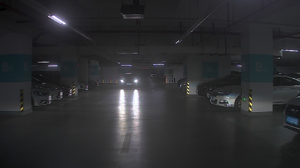}}   
    \subfigure[The texture-less wall]{
        \label{fig:c2}
        \includegraphics[width=0.31\columnwidth]{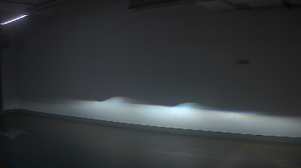}}
    \subfigure[The image blur]{
        \label{fig:c3}
        \includegraphics[width=0.31\columnwidth]{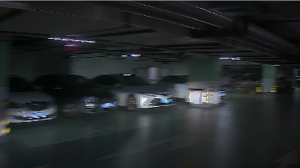}}
    \subfigure[Appearance I]{
        \label{fig:c5}
        \includegraphics[width=0.31\columnwidth]{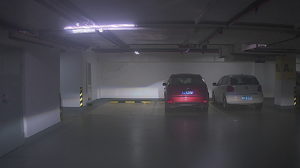}}
    \subfigure[Appearance II]{
        \label{fig:c6}
        \includegraphics[width=0.31\columnwidth]{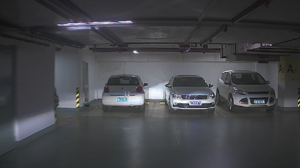}}
    \caption{Difficult cases for traditional visual SLAM methods in underground parking lots. (d) and (e) are the same place captured at different time. The appearance changes a lot.}
    \label{fig:c}
\end{figure}

This experiment was performed in an underground parking lot, where GPS is denied.
This scenario was much more challenging. 
As shown in Fig. \ref{fig:c}, there were many difficult scenes, such as the light change, the texture-less region, the motion blur, and the appearance change.
Traditional visual methods suffered from tracking lost easily.
Semantic features we used were naturally robust to these challenges.

\begin{figure}
    \centering
    \includegraphics[width=0.4\textwidth]{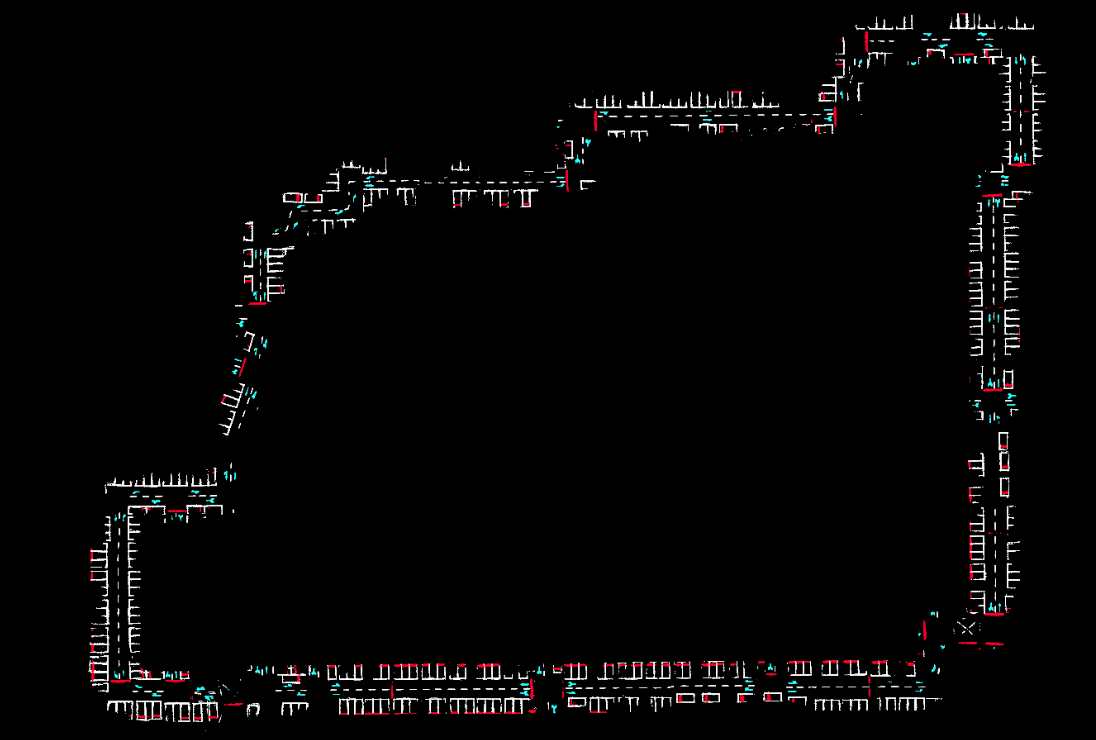}
    \caption{The semantic map used in experiment Sec. \ref{subsec:recall}.}
    \label{fig:map}
\end{figure}

We designed this experiment to validate the robustness of the proposed method for a long time.
The vehicle traversed in the parking lot several times across days.
We used the data captured on the first day to build a map.
And this map was used to localize vehicles in the following days.
The map built from AVP-SLAM is shown in Fig. \ref{fig:map}.
The parking lots was a 400m x 300m square.
We also run ORB-SLAM2 to build a feature map.
Note that ORB-SLAM2 hardly survived in this extremely difficult environment.
For a fair comparison, we used odometry (from wheel encoders and an IMU) to assist it in building the map.

\subsubsection{Recall Rate}

We counted the ratio of relocalized frames to all frames.
The relocalized frame meant the frame matched the map correctly given a prior pose.
The results of different days are shown in Table \ref{tab:Recall Rate}.
Within one hour, the recall rate of both methods was high, since the environment was almost unchanged.
With the increase of time interval, the recall rate of ORB-SLAM2 decreased dramatically.
However, the recall rate of AVP-SLAM almost remained stable.
Because ORB-SLAM2 used appearance-based features, it was easily affected by environmental change.
As shown in Fig. \ref{fig:c5} and \ref{fig:c6}, different vehicles may stay in various parking places on different days, which changed the appearance a lot.
On the contrary, semantic features were more robust than traditional features.
Even objects moved back and forth, semantic features (parking lines, guide signs, and speed bumps) always stayed there and looked the same.
They were time-invariant.
So semantic features were more suitable for dynamic environments.

\begin{table}[h]
    \centering
    \caption{{Recall Rate} \label{tab:Recall Rate}}
    \begin{tabular}{c|cc}
        \toprule
        Time Interval & ORB-SLAM2 & AVP-SLAM   \\
        \midrule
        One hour          &  77.25\% &  \textbf{82.60} \%      \\
        Three hours      & 52.67\%  &  \textbf{81.22} \%   \\
        One day        & 25.38\%     & \textbf{78.10} \%    \\
        One week        & 11.42\%   & \textbf{78.65} \%    \\
        One month        & 10.22\%   & \textbf{79.23} \%    \\
        \bottomrule
    \end{tabular}
    \begin{tablenotes}
        \footnotesize
        \item[1]  *The recall rate equals to the number of relocalized frames / the number of total frames.
    \end{tablenotes}
\end{table}

\subsubsection{Map Size}
We also compared the map size.
The result is shown in Table \ref{tab:map_size}.
The map of ORB-SLAM consisted of features and keyframes.
Each ORB feature contained descriptor in 256 bits (32 bytes), 3D position (three floats, 12 bytes), observation on the 2D image plane (two floats, 8 bytes).
The semantic map only contained the position of each feature.
Due to the view angle change, one feature in ORB-SLAM2 can be detected multiple times.
So the number of features in ORB-SLAM was more than AVP-SLAM.
The semantic map was much more efficient than the traditional descriptor map.

\begin{table}[h]
    \centering
    \caption{{Map Size} \label{tab:map_size}}
    \begin{tabular}{c|ccc}
        \toprule
        Method & Points num & Per point size & Total map size  \\
        \midrule
        ORB-SLAM2     & 647656  & 52 byte &  33.7 Mb \\
        AVP-SLAM      & 369356  & 12 byte &  4.4 Mb (574.4 kB)* \\
        \bottomrule
    \end{tabular}
    \begin{tablenotes}
        \footnotesize
        \item[1] *After Octree point cloud compression.
    \end{tablenotes}
\end{table}

\subsubsection{Localization Accuracy} 
We care about localization accuracy more than mapping accuracy in autonomous parking tasks. 
Even an inaccurate map can guide the vehicle to the parking spot as long as the vehicle can precisely localize in this map.
For metric evaluation, we manually parked the vehicle in a parking spot 20 times.
We measured the distance from the vehicle's center to the parking spot boundary in the real world.
Since the vehicle was localized on the map at the same time, we also measured this distance shown on the map.
The localization error was the difference between these two distances.
The result is shown in Table. \ref{tab:localization error}.
The average localization error is $2.36$ cm, which is sufficient for the autonomous parking task.
  
\begin{table}
    \centering
    \caption{{Localization Error} \label{tab:localization error}}
    \begin{tabular}{c|cc}
        \toprule
        Method &  Max [cm] & MEAN [cm]    \\
        \midrule
        AVP-SLAM    &  5.23  & 2.36  \\
        \bottomrule
    \end{tabular}
\end{table}

\subsection{Application: Autonomous Valet Parking}

\begin{figure}
    \centering
    \includegraphics[width=0.45\textwidth]{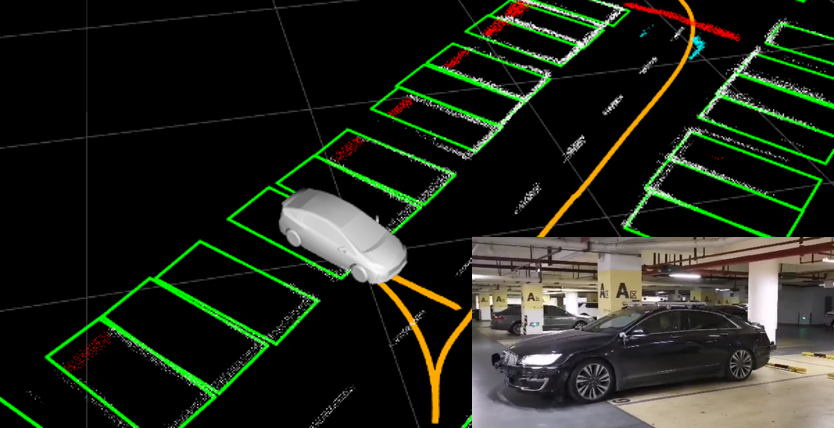}
    \caption{Autonomous Valet Parking Application.
        AVP-SLAM built a map of the parking lot in advance.
        Then this map was used to localize and navigate the vehicle going to the target parking spot autonomously. }
    \label{fig:app}
   \vspace{-0.6cm}
\end{figure}


We applied AVP-SLAM to real-world autonomous valet parking tasks, as shown in Fig \ref{fig:app}.
We used AVP-SLAM to build a map of the parking lot in advance.
Then this map was used to localize and navigate the vehicle going to the target parking spot autonomously.
Details can be found in the video material.
This experiment validated the practicability of AVP-SLAM.

\section{Conclusion}
In this paper, we proposed a visual-based localization solution, which exploits robust semantic features to assists vehicles navigating in parking lots.
Four cameras surrounding the vehicle are used.
Images are warped into a bird's eye view by IPM (Inverse Perspective Mapping).
Then the neural network detects semantic visual features which include lanes, parking lines, guide signs, and speed bumps.
A semantic visual map is built based on these features.
The vehicle can be localized in the map by semantic feature matching in centimeter-level accuracy.
The proposed system is validated by experiments and the real autonomous parking application.
AVP-SLAM achieved 1.33\% mapping error and centimeter-level localization accuracy. 

Due to the specificity of the semantic feature we used, the proposed system is only suitable for the parking lot now.
In the future, we focus on exploiting general semantic features and generalizing the proposed system for more scenarios.

\bibliography{paper.bib}

\end{document}